\pdfoutput=1
\documentclass[11pt,a4paper]{article}
\usepackage[hyperref]{acl2020}
\usepackage{times}
\usepackage{latexsym}
\usepackage{bm}
\usepackage{graphicx}
\usepackage{diagbox}

\usepackage{microtype}

\newcommand{\M}{\underline{\hskip2em}}

\aclfinalcopy 
\def\aclpaperid{1880} 

\begin{document}
		\title{ \textbf{A Two-Stage Masked LM Method for Term Set Expansion}
		}
		\author{Guy Kushilevitz \\  Technion- Israel institute of technology  \\ \texttt{guykush@cs.technion.ac.il} 
		\And
        Shaul Markovitch \\  Technion- Israel institute of technology  \\ \texttt{shaulm@cs.technion.ac.il} 
        \AND
          Yoav Goldberg \\
          Bar-Ilan University \\
          Allen Institute for AI \\
          \texttt{yogo@cs.biu.ac.il}} 

		\date{\today}
		\maketitle
\begin{abstract}
		We tackle the task of Term Set Expansion (TSE): given a small seed set of example terms from a semantic class, finding more members of that class. The task is of great practical utility, and also of theoretical utility as it requires generalization from few examples. Previous approaches to the TSE task can be characterized as either distributional or pattern-based. We harness the power of neural masked language models (MLM) and propose a novel TSE algorithm, which combines the pattern-based and distributional approaches. 
    	Due to the small size of the seed set, fine-tuning methods are not effective, calling for more creative use of the MLM.
		The gist of the idea is to use the MLM to first mine for informative patterns with respect to the seed set, and then to obtain more members of the seed class by generalizing these patterns. Our method outperforms state-of-the-art TSE algorithms. Implementation is available at: \url{https://github.com/guykush/TermSetExpansion-MPB/}
\end{abstract}
\section{Introduction}
        \textbf{Term Set expansion} (TSE)
        is the task of expanding a small seed set of terms into a larger (ideally
        complete) set of terms that belong to the same semantic category. For example,
        the seed set \textit{\{``orange'', ``apple''\}} should expand into a set of \emph{fruits},
        while \textit{\{``orange'', ``blue''\}} into a set of \emph{colors}, and
        \textit{\{``apple'',``google''\}} into a set of \emph{tech companies}. 
        Beyond being of great practical utility, the TSE task is a challenging instance
        of a \emph{generalization from few examples} problem. Solving TSE requires the
        algorithm to: (1) 
        identify the desired concept class based on few examples; and (2) 
        identify additional members of the class.
        
        We present an effective TSE method which is based on querying large, pre-trained
        masked language models (MLMs). Pre-trained language models (LMs) have been shown to
        contain semantic \cite{tenney2019bert}, syntactic \cite{DBLP:journals/corr/abs-1901-05287, structural-probe, linzen2016agreement} and factual knowledge \cite{petroni2019language}, and to be great starting
        points for transfer-learning to new tasks via fine-tuning on few
        examples. However, the TSE seed sets are too small for fine-tuning, calling for a
        different approach. Our method uses the MLMs directly for the task they were
        trained for---language-modeling---by issuing word-completion queries and operating on the
        returned word distributions.\footnote{See \cite{DBLP:conf/emnlp/AmramiG18} for a
        method that
        uses MLM word completions for word-sense induction.}
        
        \paragraph{Previous solutions} to the TSE problem (also called semantic class induction) can be roughly categorized into
        \emph{distributional} and \emph{pattern-based} approaches \cite{DBLP:conf/coling/ShiZYW10}.
        Our method can be seen as a combination of the two. 
        
        The distributional approach to TSE \cite{DBLP:conf/acl/Hindle90,
        DBLP:conf/kdd/PantelL02,DBLP:conf/emnlp/PantelCBPV09,setExpander,DBLP:conf/starsem/MahabalRM18}
        operates under the hypothesis that similar words appear in similar contexts
        \cite{DBLP:books/lib/Harris68}. 
        These methods represent each term in the vocabulary as an embedding vector that
        summarizes all the contexts the term appears in in a large corpus, and then
        look for terms with vectors that are similar to those of the seed term.
        The methods differ in their context definitions and in their way of computing
        similarities. A shortcoming of these methods is that they consider all occurrences of a term in the corpus when calculating its representation, including many contexts that are irrelevant to the concept at hand due to polysemy, noise in the corpus or non-informative contexts.
        \footnote{The work of
        \citet{DBLP:conf/starsem/MahabalRM18} is unique in this regard by considering
        only a subset of the contexts that are relevant for the expansion, as determined
        from the seed set.}
        
        In contrast, the \emph{pattern-based} approach considers \emph{specific} indicative
        patterns that signal the desired concept, looking for them in a large corpus,
        and extracting the terms that appear in them.
        Patterns can be binary 
        \cite{DBLP:conf/coling/Hearst92,DBLP:conf/wise/OhshimaOT06,DBLP:conf/acl/ZhangZSW09} (``\texttt{such as X
        or Y}''), indicating that both X and Y belong to the same class, or unary
        \cite{DBLP:conf/conll/GuptaM14,DBLP:conf/icdm/WangC07}
        (``\texttt{fruits such as X}'', ``\texttt{First I painted the wall red, but then I repainted it X}''), suggesting
        that X belongs to a certain category (\emph{fruit}, \emph{color}).
        The patterns can be determined manually \cite{DBLP:conf/coling/Hearst92} or
        automatically \cite{DBLP:conf/icdm/WangC07, DBLP:conf/conll/GuptaM14}.
        While well tailored patterns can be precise and interpretable, a notable shortcoming of pattern-based
        methods is their lack of coverage, due to the challenge of finding patterns that
        are specific enough to be accurate yet common enough in a large corpus to be
        useful. \citet{DBLP:conf/icdm/WangC07} use patterns from non-natural language
        (HTML) while \citet{DBLP:conf/conll/GuptaM14} restrict themselves to short
        patterns of 2-4 words to each side of the masked term.
        \paragraph{Our method.} 
        By using MLMs, we combine the power of the pattern-based and the distributional approaches: like the patterns-based approaches, we consider only
        specific, indicative corpus locations 
        (retaining specificity and transparency).
        We then use the distributional nature of the neural LM to generalize across patterns and corpus locations.
        
        We use sentences with a single masked location as indicative
        patterns. For example, \texttt{``We took Rexy, our pet \M, to
        the vet."} is an indicative pattern for the \emph{house animals} semantic class. 
        Given an initial set of seed terms, we first search the corpus for indicative patterns for members
        of the set (\ref{finding}). Intuitively, an indicative pattern is a corpus location which is considered by an LM to be a good fit for all seed members.
        Once we identified indicative patterns, we extend the set to
        terms that can appear in similar patterns. We propose two methods for doing this. The first method (\ref{query}) queries an MLM for completions. While effective, this method restricts the expanded set to the LM vocabulary. The second method (\ref{sim}) uses the MLM to
        define a similarity metric over patterns, and searches the corpus for terms that appear in patterns that are similar
        to the indicative ones. 
        To summarize, we embrace the pattern-based approach, while using distributional similarity for identifying good patterns as well as for generalizing across patterns.
 	
\section{Method}

\paragraph{Task formulation} we are given a seed set $S$ of $k$\footnote{In this work we focus on small values of $k$. Our experiments use $k=3$ seed terms.} terms $S=t_1,...,t_k$, that come from a larger (and unknown) gold set $S_g$. Our goal is to return  $S_g$.
Practically, our (and other) algorithms return a ranked list of terms rather than a fixed set. The evaluation is then performed over the ranking: ideally, all terms in $S_g$ will be ranked above all terms not in $S_g$.

We operate in stages. First, we search the corpus for $\ell$ indicative masked patterns $m_1,...,m_\ell$, that are likely to signal the concept class in $S_g$ with high probability. Then, we use the patterns to extend the set. 

	\subsection{Finding indicative masked-patterns} \label{finding}
    A masked pattern $m$ is a sequence of words with a single masked location (marked as ``\texttt{\M}''), where the mask indicates one or more words.
    We look for patterns 
	 such that, with high probability, instances of the desired semantic class will make good mask replacements, while instances of other classes will make bad replacements.
	 For example, ``\texttt{The capital of \M }'' is a good pattern for the ``countries'' class.
	 
	 We collect $L$ \emph{pattern candidates} for each seed term $t_j$ by querying a corpus for sentences that contain the term, and replacing the term position with a mask.
	 We then score each of the $kL$ resulting pattern candidate $m_i$, and take the $\ell$-best ones.
	 
	 Intuitively, we seek a diverse set of patterns in which all seed terms are ranked high (ie, have low rank index) in the MLM's prediction: we look for patterns whose worst-fitting seed term is still high on the list of replacement terms.
	 Formally, let $LM(m)$ be the word completions (mask replacements) predicted by the LM for pattern $m$, ranked by their probability, and let $R_{LM}(t,m)$ be the rank (index) of term $t$ in $LM(m)$.
	
	The score of the a pattern is then the maximal rank of any of the seed terms:\footnote{We assume the seed terms are a part of the LM's vocabulary.}
	\begin{equation} 
	s(m_i) = maxRank(m_i) = \max_{t_j\in S}{R_{LM}(t_j,m_i)}
	\end{equation} 
	
	We then sort the patterns by $s(m_i)$ and take the patterns with minimal values. This min-over-max formulation ensures that the patterns are a good fit for all seed terms.\footnote{Contrast this to a min-over-average formulation, which may score very well on some seed terms but badly on others.}
	
	To achieve the diversity objective, we use the following heuristic: after sorting all candidate patterns $m_i$ by $s(m_i)$, rather than taking the first $\ell$ items we go over the sorted list in order, and keep a pattern only if it differs by at least $50\%$ of it's tokens from an already kept pattern. 
	We do this until collecting $\ell$ patterns.
	
	\begin{table}[t!]
		\centering
		\begin{scalebox}{0.75}{
		\begin{tabular}{l|rrrrrr}
			\hline\hline
		\backslashbox{\textbf{\# patt}}{\textbf{\# sent}} & \textbf{20} & \textbf{100} & \textbf{300}  & \textbf{1000} & \textbf{2000} & \textbf{4000} \\  \hline
		\textbf{1} & .794 & .729 & .704 & .843 & .939 & .939 \\
		\textbf{5} & .834 & .938 & .960 & .969 & .981 & .964 \\
		\textbf{10} & .839 & .938 & .974 & .978 & .990 & .975 \\
		\textbf{20} & .838  & .932 & .972 & .987 & .990 & .978 \\
		\textbf{40} & NA  & .916 & .962 & .993 & .993 & .989 \\
		\textbf{80} & NA  & .913 & .954 & .992 & .996 & .993\\
		\textbf{160} & NA  &  NA & .949 & .985 & \textbf{.998} & .997\\
		\textbf{600} & NA  &  NA & NA & .981 & .994 & .993\\

			\hline
		\end{tabular}
		}\end{scalebox}
				\caption{
					Number of indicative patterns used ($\#patt$), and number of candidate seed-term containing sentences ($\#sent$) used for selecting these indicative patterns. Set is the NFL team set, method is MPB1. Every value is an avg MAP on 5 seeds (chosen randomly, fixed for all values of $\#sent$ and $\#patt$) of size 3. NA: $\#patt$ can not be bigger than $\#sent$. 
				}
		\label{table:paramTuning}
	\end{table}

	\subsection{seed set extension via MLM query} \label{query}
	Having identified indicative patterns, we now turn to suggest terms for expanding the seed set.
	Each indicative pattern $m_i$ naturally provides a ranked list of candidate terms $LM(m_i)=t_1,...,t_{|V|}$, where $V$ is the LM's vocabulary and each term $t_j$ is scored by its pattern-conditional probability.  We combine the term scores from all chosen indicative patterns using a product of experts approach, scoring each term by the product of probabilities (sum of log probabilities) assigned to it by each context.
	Let $p_{LM}(t|m_i)$ be the probability assigned to vocabulary term $t$ in pattern $m_i$. The term score is:
	\begin{equation}\label{eq:scoreFirst} 
    score(t) = \sum_{i=1}^\ell c_i \log p_{LM}(t|m_i) 
    \end{equation} 
	
	\noindent where $c_i = \frac{(maxRank(m_i)^{-1}}{\sum_{j=1}^\ell (maxRank(m_j)^{-1}}$ is a weighing factor for indicative pattern $m_i$, giving more weight to ``tighter'' indicative patterns. 
	
	This method is fast and effective, requiring only $\ell$ queries to the LM. However, it assumes that all the desired terms from $S_g$ appear as vocabulary items in the LM. This assumption often does not hold in practice: first, for efficiency reasons, pre-trained LM vocabularies are often small ($\sim50k$ items), precluding rare words. Second, many terms of interest are multi-word units, that do not appear as single items in the LM vocabulary.
	
	\subsection{Extended coverage via pattern similarity}
	\label{sim}
	We seek a term expansion method that will utilize the power of the pre-trained LM, without being restricted by its vocabulary: we would like to identify rare words, out-of-domain words, and multi-word units.
	
	Our solution is to generalize the indicative patterns. Rather than looking for terms that match the patterns, we instead search a large corpus for patterns which are \emph{similar} to the indicative ones, and collect the terms that appear within them. Following the distributional hypothesis, these terms should be of the desired concept class.

	By looking at patterns that surround corpus locations, we are no longer restricted by the LM vocabulary to single-token terms.
	
	However, considering all corpus locations as candidate patterns is prohibitively expensive. Instead, we take a ranking approach and restrict ourselves only to corpus locations that correspond to occurrences of candidate terms returned by a high-recall algorithm.\footnote{For example, one that is based simple distributional similarity to the seed terms. In this work we use the nearest neighbours returned by the sense2vec model \cite{DBLP:journals/corr/TraskML15}, as implemented in \url{https://spacy.io/universe/project/sense2vec}.}
	
  We use the LM to define a similarity measure between two masked patterns that aims to capture our desired notion of similarity: masked patterns are similar if they are likely to be filled by the same terms. Let $top_q(LM(m_i))$ be the $q$ highest scoring terms for pattern $m_i$. We define the similarity between two patterns as the fraction of shared terms in their top $q$ predictions ($q$ being a hyperparameter): \\[0.5em]
 $sim(m_i,m_j)=$\\
 \indent\indent\indent$|top_q(LM(m_i))\cap top_q(LM(m_j))|/q$
 
For a candidate term $t$, let $pats(t)=m^t_1,...,m^t_n$ be the set of patterns derived from it: sentences that contain $t$, where $t$ is replaced with a mask. Note that $t$ can be an arbitrary word or word sequence.
We wish to find terms for which the similarity between $pats(t)$ and the indicative patterns is high. However, since words have different senses, it is sufficient for only \emph{some} patterns in $pats(t)$ to be similar to patterns in $m_1,...,m_\ell$. 
We score a term $t$ as:
	\begin{equation}\label{eq:similarityScore} 
    score(t) = \sum_{i=1}^\ell c_i  \smash{\displaystyle\max_{m \in pats(t)} sim(m_i, m)}
    \end{equation} 

\noindent where $c_i$ is the pattern weighing factor from equation (2). As $ \sum_{i=1}^\ell c_i = 1 $, the term score $score(t)$ for every term t is $\in [0,1]$.
	\begin{table}[t!]
		\centering
		\begin{scalebox}{0.75}{
		\begin{tabular}{l|rrrrrr}
			\hline\hline
		\textbf{Set} & \textbf{k=1} & \textbf{k=5} & \textbf{k=50} & \textbf{k=300} &\textbf{ k=700} & \textbf{k=3000} \\  \hline
		\textbf{States} & .693 & .848 & .986 & .965 & .972 & .975\\
		\textbf{NFL} & .876 & .939 & .938 & .919 & .921 & .916\\
			\hline
		\end{tabular}
		}\end{scalebox}
				\caption{Effect of similarity measure's $k$ on performance, using MPB2 on a single random seed from each set.}
		\label{table:Similarity}
	\end{table}
\section{Experiments and Results}
    We refer to the method in Section (2.2) as \textbf{MPB1} and the method in section (2.3) as \textbf{MPB2}. \\
	\noindent\textbf{Setup.} 
	In our experiments we use BERT \cite{Bert} as the MLM, and English Wikipedia as the corpus. Following previous TSE work (e.g. \cite{DBLP:conf/starsem/MahabalRM18}), we measure performance using MAP (using MAP$_{70}$ for the open set).
	For each method we report the average MAP over several runs (exact number mentioned under each table), each with a different random seed set of size 3.
	Based on preliminary experiments, for MPB1 we use $\ell=160$ and $L=2000/k$ and for MPB2 we use $\ell=20$ and $L=2000/k$. \footnote{see Additional experiments for justification of these parameter choices}
	When comparing different systems (i.e, in Table \ref{table:MapScores}), each system sees the same random seed sets as the others. For smaller sets we expand to a set of size 200, while for the Countries and Capitals sets, which have expected sizes of $>100$, we expand to 350 items.
	
	\noindent\textbf{Dataset.} 
	Automatic TSE evaluation is challenging. A good TSE evaluation set should be \emph{complete} (contain all terms in the semantic class), \emph{clean} (not contain other terms) and \emph{comprehensive} (contain all different synonyms for all terms). These are hard to come by. Indeed, previous work either used a small number of sets, or used some automatic set acquiring method which commonly are not complete. We curated a dataset with 7 closed, well defined sets, which we make publicly available. The sets are National football league teams (NFL, size:32), Major league baseball teams (MLB, 30), US states (States, 50), Countries (Cntrs, 195), European countries (Euro, 44) Capital cities (Caps, 195) and Presidents of the USA (Pres, 44). We also provide on one open class set: Music Genres (Genre). This set created by manually verifying the items in the union of the output of all the different algorithms. This set contains around 600 unique items.\\
	\noindent\textbf{Compared Methods.}
  We compare our methods, \textsc{Mpb1} (MLM-pattern-based) (Section \ref{query}) and \textsc{Mpb2}\footnote{We follow \cite{DBLP:conf/starsem/MahabalRM18} and limit MPB2 to 200,000 most frequent terms. MPB2 can work with any number of terms and is limited only by the candidate supplying method (in this implementation- sence2vec which has $\sim$3,400,000 terms).} (Section \ref{sim}), to two state-of-the-art systems: setExpander\footnote{We use the non-grouping release version because it reaches better results on our dataset than the grouping one.} (\textsc{se}) \cite{setExpander}, and category builder (\textsc{cb}) \cite{DBLP:conf/starsem/MahabalRM18}.  
  We also compare to two baselines: The first, \textsc{bb} (basic-BERT), is a baseline for MPB1. This is a BERT-based baseline that uses the \textsc{Mpb1} method on patterns derived from sentences that include seed terms, without the selection method described in Section \ref{finding}. The second, \textsc{s2v}, is a baseline for MPB2. This is a basic distributional method that uses sense2vec \cite{DBLP:journals/corr/TraskML15} representations,\footnote{\url{https://explosion.ai/demos/sense2vec}} which is also our candidate acquisition method for \textsc{Mpb2} (\ref{cand}). 
  As \textsc{Mpb2} relies on external candidate generation, we also report on the oracle case \textsc{Mpb2+O} where we expand the \textsc{s2v}-generated candidate list to include all the members of the class.\\

	\begin{table*}[ht!]
		\centering
		\begin{scalebox}{0.91}{
		\begin{tabular}{ll|rrrrrrr|r||r}
			\hline\hline
			\textbf{Method} && \textbf{NFL} & \textbf{MLB} & \textbf{Pres} & \textbf{States}  & \textbf{Cntrs}  & \textbf{Euro} &  \textbf{Caps} & \textbf{Genre} & \textbf{Avg} \\  \hline
			SE & (SetExpander) &.54 & .45 & .33 & .55 & .55 & .61 & .14 & \textbf{.99} & .52 \\ 
			CB & (CategoryBuilder) &  \textbf{.98} &  .97 & \textbf{.70} & .93 & .74 & .46 & .21 & .67 &.71 \\ 
			BB & (BERT Baseline) & .91 & .92* & .52** & NA & NA  & NA & NA  & NA & .78$\dagger$ \\
			\textbf{MPB1} & (Section 2.2) & \textbf{.98} & \textbf{.99} * & .63** & NA & NA  & NA & NA & NA & .87$\dagger$ \\
			S2V & (Sense2Vec Baseline) & .95 & .80 & .18 & .94 & .71 & .78 & .21 & .90 & .68\\ 
			\textbf{MPB2} & (Section 2.3) & .95 & .82 &.37 & \textbf{.98} & \textbf{.76} & \textbf{.79} & \textbf{.27} & .98 & .74\\ 
			\hline
			\textbf{MPB2+O} & (Sec 2.3, Oracle) & .95 &.90 &.88 &.98 & .91 & .81 & .80 & NA' & .89$\dagger$ \\ 
			\hline
		\end{tabular}
		}\end{scalebox}
				\caption{
				Main results. Average MAP scores over 3 random seeds of size 3. \textbf{*/**}: excluding 2 or 3 OOV terms. \textbf{NA}: Not applicable, because sets contain many OOV terms. \textbf{NA'}: Not applicable for oracle setting, because gold standard candidates not available for open sets. $\bm{\dagger}$: Average value over applicable sets only.
				}
		\label{table:MapScores}
	\end{table*}
	
\noindent\textbf{Main Results.}
Our main results are reported in Table \ref{table:MapScores}. 
Our first method, MPB1, achieves the best scores on two of the three sets suitable for its limitations (where all or most of the set's terms are in the LM's vocabulary), and second-best results on the third.\footnote{MPB1's relatively poor performance on the president's set can be a result of the basic terms MPB1 considers. MPB1 ranks only terms which are in the LM's vocabulary, which means that while other expanders can rank terms like "President George W. Bush", MPB1 will consider terms like "bush", which are harder to ascribe to the presidents set. While this is true for all sets, it seams to be more significant for a set containing person names.} MPB2 outperforms all other methods on 5 out of 7 closed sets when assuming gold-standard candidates (MPB2+O), and even when considering the missing candidates it outperforms other expanders on 4 out of 7 closed sets, averaging the best MAP score on all sets. While other methods tend to stand out in either closed sets (CB) or the open set (SE),\footnote{SE does not rank the seed terms, as opposed to other methods. For fairness, we add them in the beginning of the returned list before computing the MAP score.} MPB2 shows good performance on both kinds of sets. The results also suggest that a better candidate-acquiring method may lead to even better performance. 

\paragraph{Additional experiments.}
\label{exp}
\emph{How many sentences should we query when searching for indicative patterns, and how many patterns should we retain?} Table \ref{table:paramTuning} shows a grid of these parameters. We use the NFL set for this experiment, as terms in this set all have more than one meaning, and for most the common usage is not the one that belongs to the NFL set (e.g \textit{"jets", "dolphins"}). Therefore, this set should give a pessimistic estimation for the the number of sentences we need to extract to find quality indicative patterns. Results imply that $\sim$2000 appearances of seed terms are sufficient, and that good results can be obtained also with fewer instances. This shows that---beyond the data used to train the initial MLM---we do not require a large corpus to achieve good results, suggesting applicability also in new domains.\footnote{While for MPB1 there are no prominent downsides in using a large number of indicative patterns, for MPB2 doing so will force us to use a large number of occurrences of the candidate terms also. This will (1) be costly run-time wise and (2) many occurrences of rare terms might not always be available. Therefore, we choose different parameters for MPB1 and MPB2. While in both we will use 2000 sentences to search for these indicative patterns ($L=2000/k$), for MPB1 we will use 160 indicative patterns ($\ell=160$) and for MPB2 we will use only 20 of them ($\ell=20$).}

\emph{How sensitive is the algorithm to the choice of $k$ when computing the pattern similarity?} Table \ref{table:Similarity} shows that the similarity measure is effective for various $k$ values, with max performance at $\sim$50. 

Finally, \emph{how do the different methods behave in a case where the seed terms are a part of a subset?} Table \ref{table:SetSubset} shows a case where seed terms are European countries. Ideally, we would like top results to be European countries, later results to be non-European countries, and then unrelated terms. MPB2+O achieves the best MAP scores on both the set and the subset. In the subset case, even when not provided with all oracle terms, MPB2 is better then all other expanders. While other expanders tend to reach stronger results on either the set or the subset, MPB2+O achieves similar scores on both. 

	\begin{table}[t!]
		\centering
		\begin{scalebox}{0.75}{
		\begin{tabular}{l|rrr|rr}
			\hline\hline
			\textbf{Set} & \textbf{S2V} & \textbf{CB}  & \textbf{SE}  & \textbf{MPB2} &  \textbf{MPB2+O}\\  \hline
			\textbf{European countries} & .782 & .458 & .609 & .787 & .814\\
		    \textbf{Countries} & .454 & .752 & .197 & .528 & .804\\ 
			\hline
		\end{tabular}
		}\end{scalebox}
				\caption{
					Performance on a subset. Avg MAP over 3 random seeds of size 3.
				}
		\label{table:SetSubset}
	\end{table}
	\section{Conclusions}
	We introduce an LM-based TSE method, reaching state-of-the-art results. The method uses the power of LM predictions to locate indicative patterns for the concept class indicated by the seed terms, and then to generalize these patterns to other corpus locations. Beyond strong TSE results, our method demonstrates a novel use of pre-trained MLMs, using their predictions directly rather than relying on their states for fine-tuning.

\section*{Acknowledgements}
This project has received funding from the European Research Council (ERC) under the European Union's Horizon 2020 research and innovation programme, grant agreement No. 802774 (iEXTRACT).
	
    \bibliographystyle{acl_natbib}
    \bibliography{acl2020}
    
\section{Appendix 1: Finding candidate terms} \label{cand}
	For our first method, MPB1, the candidate terms we score are just the terms in the LM's vocabulary. For our second method, MPB2, we want to score candidates which are not in this vocabulary as well. Hence, we need a way to acquire these candidates. As running on all possible terms is prohibitive, we seek an efficient method to acquire a high-recall group of candidates for the desired semantic class. We get this using a simple distributional set-expander: we compute the mean vector for words in our seed set, and look for the top-k neighbours in a distributional space.
	
	Specifically, we use the sense2vec pretrained vectors.
	Sense2vec \cite{DBLP:journals/corr/TraskML15} is a misleadingly-named algorithm from the w2v-family \cite{DBLP:journals/corr/abs-1301-3781} that models each term as \textit{``term|part of speech"}. This allows it, for example, to learn different representations for \textit{``duck|verb"} and \textit{``duck|noun"}. 
	
	More importantly, the pre-trained sense2vec vectors distributed by explosion.ai\footnote{\url{https://explosion.ai/demos/sense2vec}} are trained over a large and diverse English corpus (reddit posts and comments from 2015 and 2019),
	and its vocabulary includes not only single words but also multi-word units (NP-chunks and named entities).
\end{document}